\documentclass[letterpaper, 10 pt, conference]{ieeeconf}  

\IEEEoverridecommandlockouts                              

\usepackage[T1]{fontenc}
\usepackage{cite}
\usepackage{amssymb,amsfonts}
\usepackage{blindtext}
\makeatletter
\let\NAT@parse\undefined
\makeatother
\usepackage{hyperref}
\usepackage{graphicx}
\usepackage{textcomp}
\usepackage{mathtools}
\usepackage{changes}
\usepackage{subcaption}
\usepackage{algorithm}
\usepackage{xcolor}
\usepackage{color, soul}
\usepackage{amsmath}
\usepackage{booktabs}
\usepackage{multirow}
\usepackage{xstring}
\usepackage{hhline}
\usepackage{kotex}
\usepackage{siunitx}
\usepackage{comment}
\usepackage{physics}
\usepackage{tikz}
\usepackage{cleveref}
\usepackage{lipsum}
\usepackage{bm}
\usepackage{siunitx}
\usepackage{array}
\usepackage{eqparbox}
\usepackage{caption}
\usepackage[noend]{algorithmic}

\usepackage{etoolbox}  
\makeatletter
\patchcmd{\algorithmic}{\addtolength{\ALC@tlm}{\leftmargin} }{\addtolength{\ALC@tlm}{\leftmargin}}{}{}
\makeatother

\crefformat{section}{\S#2#1#3} 
\crefformat{subsection}{\S#2#1#3}
\crefformat{subsubsection}{\S#2#1#3}

\floatname{algorithm}{Algorithm}

\definecolor{rv}{RGB}{0, 0, 0}

\sethlcolor{lightgray}

\makeatletter
\DeclareRobustCommand{\iscircle}{\mathord{\mathpalette\is@circle\relax}}
\newcommand\is@circle[2]{%
  \begingroup
  \sbox\z@{\raisebox{\depth}{$\m@th#1\bigcirc$}}%
  \sbox\tw@{$#1\square$}%
  \resizebox{!}{\ht\tw@}{\usebox{\z@}}%
  \endgroup
}
\makeatother

\IEEEoverridecommandlockouts                              

\usepackage{amsmath} 
\usepackage{caption}
\newcommand{\rom}[1]{\uppercase\expandafter{\romannumeral #1\relax}}

\title{\LARGE \bf B-TMS: Bayesian Traversable Terrain Modeling and Segmentation Across 3D LiDAR Scans and Maps for Enhanced Off-Road Navigation}

\author{
    Minho Oh$^{1}$, \textit{Student Member, IEEE}, Gunhee Shin$^{1}$, Seoyeon Jang$^{1}$, Seungjae Lee$^{1}$, Dongkyu Lee$^{1}$, \\ Wonho Song$^{1}$, Byeongho Yu$^{1}$, Hyungtae Lim$^{1}$, Jaeyoung Lee$^{2}$, and Hyun Myung$^{1,*}$, \textit{Senior Member, IEEE}
    \thanks{
        This study was funded and supported by the grants from Hanwha Aerospace as part of the development of autonomous driving technology for unstructured environment. The students are supported by BK21 FOUR.
    }
    \thanks{$^*$Corresponding author: Hyun Myung}
    \thanks{
        \indent $^1$The authors are with the School of Electrical Engineering and KI-R at Korea Advanced Institute of Science and Technology (KAIST), Daejeon, 34141, Republic of Korea. 
        {\tt\footnotesize 
        \{minho.oh, gunhee$\_$shin, 9uantum01, sj98lee, dklee, swh4613, bhyu, shapelim, hmyung\}@kaist.ac.kr}. 
        \indent $^{2}$The author is with Hanwha Aerospace, 6, Pangyo-ro 319 beon-gil, Bundang-gu, Seongnam-si, Gyeonggi-do, Republic of Korea.
        {\tt\footnotesize \{jaeyoung1.lee\}@hanwha.com} \hfill \break
    }
}

\begin{document}

\captionsetup[figure]{labelformat={default},labelsep=period,name={Fig.}}


\maketitle
\thispagestyle{empty}
\pagestyle{empty}

\begin{abstract}
Recognizing traversable terrain from 3D point cloud data is critical, as it directly impacts the performance of autonomous navigation in off-road environments. 
However, existing segmentation algorithms often struggle with challenges related to changes in data distribution, environmental specificity, and sensor variations. 
Moreover, when encountering sunken areas, their performance is frequently compromised, and they may even fail to recognize them. 
To address these challenges, we introduce \textit{B-TMS}, a novel approach that performs map-wise terrain modeling and segmentation by utilizing Bayesian generalized kernel (BGK) within the graph structure known as the tri-grid field (TGF). 
Our experiments encompass various data distributions, ranging from single scans to partial maps, utilizing both public datasets representing urban scenes and off-road environments, and our own dataset acquired from extremely bumpy terrains. 
Our results demonstrate notable contributions, particularly in terms of robustness to data distribution variations, adaptability to diverse environmental conditions, and resilience against the challenges associated with parameter changes.
\end{abstract}


\vspace{-2mm}
\section{Introduction}
\vspace{-2mm}

In the field of robotics, there is a growing demand for the recognition and accurate representation of the surrounding environment. 
In particular, recognizing terrain data for unmanned ground vehicles (UGVs) has become increasingly important~\cite{lim24simbutdiff}. 
Numerous research efforts have been concentrated on enhancing drivable region detection, object identification~\cite{zermas2017fast, xue2021lidar, oh22travel}, static map generation~\cite{lim21erasor, lim23erasor2, jang23toss}, labeling dynamic objects~\cite{chen22automos}, odometry estimation~\cite{shan2018lego, seo22pagoloam, song23bigstep}, and global localization~\cite{lim2023quatro++} by utilizing terrain estimation. 
However, the off-road terrain recognition, which encompasses diverse and uneven landscapes, still remains a formidable challenge.

Existing ground segmentation methods primarily focus on flat urban scenes~\cite{fischler1981ransac, moosmann2009segmentation, zermas2017fast}. 
Xue~\textit{et al.} introduced a drivable terrain detection method that employs edge detection in normal maps to segment areas between curbs or walls~\cite{xue2021lidar}. 
Addressing non-flat and sloped terrains, Narksri \textit{et al.} proposed a multi-region RANSAC plane fitting approach~\cite{narksri2018slope}. 
Wen~\textit{et al.} utilized LiDAR range- and $z$-images that combines features with different receptive field sizes to improve ground recognition~\cite{wen2023dipgseg}. 
Paigwar~\textit{et al.} put forth a learning-based terrain elevation representation~\cite{paigwar2020gndnet}.
However, these existing methods face challenges when applied to off-road and irregular bumpy terrain.

Our prior work has been primarily centered on enhancing off-road autonomous driving performance. 
Initially, we proposed a PCA-based multi-section ground plane fitting algorithm~\cite{lim21patchwork}, and subsequently improved its robustness against outliers frequently encountered in 3D LiDAR data~\cite{lee2022patchwork++}. 
We also introduced a graph-based traversability-aware approach~\cite{oh22travel}. 
Despite our efforts to enhance ground segmentation in off-road environments such as forested areas, our previous approaches still face challenges, including the need for parameter adjustments based on data distribution and difficulties in recognizing unobservable or sunken areas.

\begin{figure}[t]
    \captionsetup{font=footnotesize}
    \centering
    \includegraphics[width=\columnwidth]{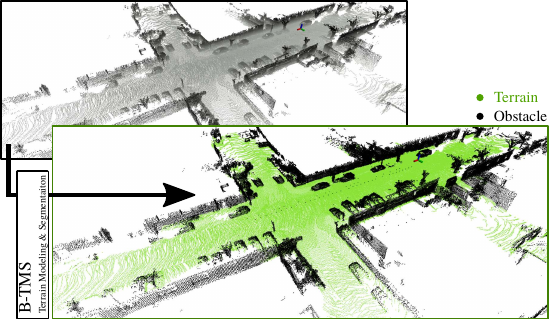}
    \caption{Overview of map-wise Bayesian-based traversable terrain modeling and segmentation (B-TMS). B-TMS models and segments traversable terrain data in the given 3D point cloud map at once.}
    \label{fig:main_fig}
    \vspace{-5mm}
\end{figure}

In this study, by extending our previous research~\cite{oh22travel}, we introduce B-TMS, a novel approach for integrating probability approach with tri-grid field (TGF)-based terrain modeling and analyzing map-wise traversable terrain regions, as illustrated in Fig.~\ref{fig:main_fig}. We have overcome the limitations of existing methods and conducted evaluation across three diverse datasets, demonstrating the following contributions:
\begin{itemize}
    \item This research marks the pioneering map-wise terrain segmentation, exhibiting robustness against changes in data distribution stemming from map scale changes, for example.
    \item Integration of BGK-based terrain model completion with our global TGF has significantly reduced the performance change gap owing to the parameter alterations.
    \item Environmental adaptability is proved through evaluations in both urban and off-road environments, as well as in extremely bumpy terrain scenarios. 
\end{itemize}


\section{Terrain Modeling and Segmentation}
B-TMS mainly consists of initial traversable terrain search on global TGF with breadth-first traversable graph search (B-TGS), BGK-based terrain model completion, and traversability-aware global terrain model fitting modules.

\subsection{Initial Traversable Terrain Search on Global TGF}
Firstly, as proposed in our previous work~\cite{oh22travel}, we form the global graph structure known as the global TGF as follows:
\begin{equation}
    \mathbf{N}^{\mathcal{T}}=\{\mathbf{n}^{\mathcal{T}}_{i}|i\in\mathcal{N}\}, 
    \mathbf{E}^{\mathcal{T}}=\{\mathbf{e}^{\mathcal{T}}_{ij}|i,j\in\mathcal{N}\},
\end{equation}
where $\mathbf{N}^{\mathcal{T}}$, $\mathbf{E}^{\mathcal{T}}$, and $\mathcal{N}$ represent a set of nodes $\mathbf{n}^{\mathcal{T}}_{i}$ whose center location is defined as $\mathbf{x}_{i}\in \Bbb{R}^2$, a set of edges $\mathbf{e}^{\mathcal{T}}_{ij}$, and the total number of nodes, respectively. 
3D cloud data is embedded into TGF by global $xy$-coordinate location with a resolution $r^{\mathcal{T}}$, then each $\mathbf{n}^{\mathcal{T}}_{i}$ contains the corresponding points $\mathcal{P}_i$. And by applying PCA-based plane fitting to $\mathcal{P}_i$, the planar model $\mathbf{P}_i$ of $\mathbf{n}^{\mathcal{T}}_{i}$ can be initially defined as follows:
\begin{equation}
\begin{split}
    \mathbf{P}_{i}^{\mathsf{T}} 
    \begin{bmatrix}
        \mathbf{m}_{i}\\
        1
    \end{bmatrix} 
    &= 
    \begin{bmatrix}
    \mathbf{s}_{i}^{\mathsf{T}} & d_{i}
    \end{bmatrix}
    \begin{bmatrix}
        \mathbf{m}_{i}\\
        1
    \end{bmatrix}
    = 0,
\end{split}
\label{eq:planar_model}
\end{equation}
where $\mathbf{m}$, $\mathbf{s}$, and $d$ represent the mean point, surface normal vector, and plane coefficient, respectively.

Additionally, with the obtained descending ordered eigenvalues $\lambda_{k\in{1,2,3}}$, the traversability weight $\bar{w}^{\mathcal{T}}_i$ is calculated as follows:
\begin{equation}
    \bar{w}^{\mathcal{T}}_{i} = (1 - \lambda_{3,i}/\lambda_{1,i}) \cdot ((\lambda_{2,i}-\lambda_{3,i})/\lambda_{1,i}) \in[0,1].
\end{equation}
Please note that to facilitate BGK-based terrain model and to obtain a normalized weight, $\bar{w}^{\mathcal{T}}$ is defined with scattering, $\lambda_{3}/\lambda_{1}$, and planarity, $(\lambda_{2}-\lambda_{3})/\lambda_{1}$, as defined in Weinmann~\textit{et al.}~\cite{weinmann2015semantic}, which is different from~\cite{oh22travel}.
So each node in the global TGF can be expressed as follows:
\begin{equation}
    \mathbf{n}^{\mathcal{T}}_{i} = \{\mathbf{x}_{i}, \mathcal{P}_i, \mathbf{m}_{i}, \mathbf{s}_{i}^{\mathsf{T}}, d_{i}, \bar{w}^{\mathcal{T}}_{i}\} \in\mathbf{N}^{\mathcal{T}}.
\end{equation}
Then, to classify the initial terrain nodes, each node is classified into terrain node $\mathbf{n}^{\mathcal{T},t}$ and others $\mathbf{n}^{\mathcal{T},o}$ by the inclination threshold, $\theta^{\mathcal{T}}$, and the threshold $\sigma^{\mathcal{T}}$ for the number of $\mathcal{P}_i$ as follows:
\begin{equation}
    \mathbf{n}^{\mathcal{T}}_{i} \Rightarrow
    \begin{cases}
        \mathbf{n}^{\mathcal{T},t}_{i}, & \text{if } \cos(z_{\mathbf{s}^{\mathsf{T}}_{i}}) \geq \cos(\theta^{\mathcal{T}})\wedge n(\mathcal{P}_i) \leq \sigma^{\mathcal{T}}\\
        \mathbf{n}^{\mathcal{T},o}_{i}, & \text{otherwise} \\
    \end{cases},
\end{equation}
where $z_{\mathbf{s}^{\mathsf{T}}_{i}}$ is a $z$-axis component of $\mathbf{s}^{\mathsf{T}}_{i}$.

To search for a set of traversable nodes in the global TGF, we adopt the B-TGS approach based on $lcc(\cdot)$ which determines the local convexity and concavity~\cite{oh22travel}.
$lcc(\mathbf{e}^{\mathcal{T},t}_{ij})$ confirms the local traversability between $\mathbf{n}^{\mathcal{T},t}_{i}$ and $\mathbf{n}^{\mathcal{T},t}_{j}$ as follows:
\begin{equation}
lcc(\mathbf{e}^{\mathcal{T}}_{ij}) = \begin{cases} 
    true,  & \text{if } |\mathbf{s}_i\cdot\mathbf{s}_j|     \geq 1 - \sin(||\mathbf{d}_{ij}||\epsilon_2)   \\ 
    & \wedge\:          |\mathbf{s}_j\cdot\mathbf{d}_{ji}|  \leq ||\mathbf{d}_{ji}||\sin\epsilon_1        \\
    & \wedge\:          |\mathbf{s}_i\cdot\mathbf{d}_{ij}|  \leq ||\mathbf{d}_{ij}||\sin\epsilon_1        \\
    false, & \text{otherwise},
\end{cases}
\label{eq:lcc}
\end{equation}
where \(\mathbf{d}_{ji}=\mathbf{m}_i-\mathbf{m}_j\) is the displacement vector. \(\epsilon_1\) and \(\epsilon_2\) denote the thresholds regarding normal similarity and plane convexity, respectively.
As a result of the B-TGS process, only the searched traversable terrain nodes remain classified as $\mathbf{n}^{\mathcal{T},t}$, while the others are reclassified as $\mathbf{n}^{\mathcal{T},o}$.

\subsection{BGK-based Terrain Model Completion}
In the terrain model completion module, the terrain planar models of $\mathbf{n}^{\mathcal{T},o}$ are predicted using the remaining $\mathbf{n}^{\mathcal{T},t}$. 
For the neighbor-based prediction, we propose the BGK-based terrain model prediction method on global TGF.
Therefore, before predicting the terrain model of $\mathbf{n}^{\mathcal{T},o}_{j}$, we utilize the BGK function $k(\cdot,\cdot)$ which estimates the likelihood of it being influenced by $\mathbf{n}^{\mathcal{T},t}_{i}$, inspired by~\cite{melkumyan2009sparse} as follows:
\begin{multline}
    k(\mathbf{n}^{\mathcal{T},t}_{i},\mathbf{n}^{\mathcal{T},o}_{j})=
    \\
    \begin{cases}
    \frac{(2+\cos(2\pi \frac{d_{ij}}{l}))(1-\frac{d_{ij}}{l})}{3}+\frac{\sin(2\pi \frac{d_{ij}}{l})}{2\pi}, &\mathrm{if}~\frac{d_{ij}}{l} < 1
    \\
    0 , &\mathrm{otherwise}
    \end{cases}
\end{multline}
where $d_{ij}$ is the 2D $xy$-distance between $\mathbf{m}_{i}$ and $\mathbf{x}_{j}$ and $l$ is the radius of the prediction kernel $\mathcal{K}^{\mathcal{T}}_{j}$. 
Under the assumption that the $xy$-coordinates between $\mathbf{m}_{j}$ and $\mathbf{x}_{i}$ of $\mathbf{n}^{\mathcal{T},o}_{j}$ are the same, the $z$-value of $\mathbf{m}_{j}$ can be easily predicted as follows:
\begin{equation}
    \mathcal{L}_{z}(\mathcal{K}^{\mathcal{T}}_{j})\triangleq z_{j} 
    = \frac{\sum_{\mathbf{n}^{\mathcal{T}}_i}^{\mathcal{K}^{\mathcal{T}}_{j}} k(\mathbf{n}^{\mathcal{T},t}_{i},\mathbf{n}^{\mathcal{T},o}_{j})\cdot z_{i}}
           {\sum_{\mathbf{n}^{\mathcal{T}}_i}^{\mathcal{K}^{\mathcal{T}}_{j}} k(\mathbf{n}^{\mathcal{T},t}_{i},\mathbf{n}^{\mathcal{T},o}_{j})},
    \label{eq:z_inference}
\end{equation}
where $\mathcal{L}_{z}(\cdot)$ denotes the inference function of $z$.

Furthermore, to predict $\mathbf{s}_{j}$, we set the assumption that $\mathbf{s}_{j}$ is perpendicular to $\Delta = \mathbf{m}_{j}-\mathbf{m}_{i}$. 
So, we can model the normal vector of $\mathbf{n}^{\mathcal{T}}_{j}$ affected by $\mathbf{n}^{\mathcal{T}}_{i}$, $\mathbf{s}_{j\leftarrow i}$ as (\ref{eq:normal_model}), and $\mathbf{s}_{j}$ can also be predicted by the inference function as (\ref{eq:normal_inference}).
\begin{equation}
    \mathbf{s}_{j\leftarrow i}^{\mathsf{T}} = \frac{1}{||\Delta||}[\frac{-\Delta_x\Delta_z}{\sqrt{\Delta_{x}^{2}+\Delta_{y}^{2}}},\frac{-\Delta_y\Delta_z}{\sqrt{\Delta_{x}^{2}+\Delta_{y}^{2}}}, {\sqrt{\Delta_{x}^{2}+\Delta_{y}^{2}}}]
    \label{eq:normal_model}
\end{equation}
\begin{equation}
    \mathcal{L}_{\mathbf{s}}(\mathcal{K}^{\mathcal{T}}_{j}) \triangleq \mathbf{s}_{j}
    = \frac{\sum_{\mathbf{n}^{\mathcal{T}}_i}^{\mathcal{K}^{\mathcal{T}}_{j}} k(\mathbf{n}^{\mathcal{T}}_{i},\mathbf{n}^{\mathcal{T}}_{j})\cdot \mathbf{s}_{j\leftarrow i}}
           {\sum_{\mathbf{n}^{\mathcal{T}}_i}^{\mathcal{K}^{\mathcal{T}}_{j}} k(\mathbf{n}^{\mathcal{T}}_{i},\mathbf{n}^{\mathcal{T}}_{j})}
    \label{eq:normal_inference}
\end{equation}
where $\mathcal{L}_{\mathbf{s}}(\cdot)$ denotes the inference function of $\mathbf{s}$.

The plane coefficient, $d_{j}$, can be estimated by (\ref{eq:planar_model}).
Lastly, for prediction of $\bar{w}_{j}^{\mathcal{T}}$, we define the inference function $\mathcal{L}_{w}(\cdot)$ as follows:
\begin{equation}
    \mathcal{L}_{w}(\mathcal{K}^{\mathcal{T}}_{j}) \triangleq
    \bar{w}_{j}^{\mathcal{T}}
    = \frac{\sum_{\mathbf{n}^{\mathcal{T}}_i}^{\mathcal{K}^{\mathcal{T}}_{j}} k(\mathbf{n}^{\mathcal{T},t}_{i},\mathbf{n}^{\mathcal{T},o}_{j})\cdot \bar{w}_{i}^{\mathcal{T}}(\mathbf{s}_{i}\cdot\mathbf{s}_{j})}
           {\sum_{\mathbf{n}^{\mathcal{T},t}_i}^{\mathcal{K}^{\mathcal{T},o}_{j}} k(\mathbf{n}^{\mathcal{T},t}_{i},\mathbf{n}^{\mathcal{T},o}_{j})},
    \label{eq:weight_inference}
\end{equation}
considering the similarity of normal vectors. This is because traversability is related to the similarity to existing terrain models. 
By utilizing our proposed BGK-based terrain model prediction on global TGF, some $\mathbf{n}^{\mathcal{T},o}$ are reverted to $\mathbf{n}^{\mathcal{T},t}$.

\subsection{Traversability-aware Global Terrain Model Fitting}

Finally, in this traverasbility-aware global terrain process, every $\mathbf{n}^{\mathcal{T}}_i \in \mathbf{N}^{\mathcal{T}}$ are updated as $\hat{\mathbf{n}}^{\mathcal{T}}_i \in \mathbf{N}^{\mathcal{T}}$. 
So, by applying weighted corner fitting approach to all tri-grid corners, which was proposed in our previous work~\cite{oh22travel},
$\mathbf{n}^{\mathcal{T}}\in\mathbf{N}^{\mathcal{T}}$, which are surrounded by three weighted corners $\hat{\mathbf{c}}_{m\in{1,2,3}}\in\Bbb{R}^3$, are updated as follows:
\begin{equation}
    \hat{\mathbf{n}}^{\mathcal{T}}_{i} = \{\mathbf{x}_{i}, \mathcal{P}_i, \hat{\mathbf{m}}_{i}, \hat{\mathbf{s}}_{i}^{\mathsf{T}}, \hat{d}_{i}, \bar{w}^{\mathcal{T}}_{i}\} \in\mathbf{N}^{\mathcal{T}},
\end{equation}
\begin{equation}
\begin{split}
    \hat{\mathbf{P}}_{i} = 
        \begin{bmatrix}
            \hat{\mathbf{s}}_{i} & \hat{d}_{i}
        \end{bmatrix}
    &,\:\: \hat{\mathbf{m}}_{i} = (\hat{\mathbf{c}}_{i,1}+\hat{\mathbf{c}}_{i,2}+\hat{\mathbf{c}}_{i,3})/3,
    \\
    \hat{d}_{i} = - \hat{\mathbf{s}}_{i} \cdot \hat{\mathbf{m}}_{i}
    &,\:\:
    \hat{\mathbf{s}}_{i} =\frac{(\hat{\mathbf{c}}_{i,2}-\hat{\mathbf{c}}_{i,1})}{||\hat{\mathbf{c}}_{i,2}-\hat{\mathbf{c}}_{i,1}||} 
                      \cross 
                      \frac{(\hat{\mathbf{c}}_{i,3}-\hat{\mathbf{c}}_{i,1})}{||\hat{\mathbf{c}}_{i,3}-\hat{\mathbf{c}}_{i,1}||}.
\end{split}
\end{equation}

Finally, based on the updated nodes $\hat{\mathbf{n}}^{\mathcal{T}}_{i}$ in global TGF, each point $\mathbf{p}_{k} \in \mathcal{P}_i$ is segmented as follows:
\begin{equation}
\begin{split}
        \texttt{label}(\mathbf{p}_{k})&=
        \begin{cases}
        \texttt{Terrain},& \text{if }\mathbf{p}_{k}\cdot\hat{\mathbf{s}}_i+\hat{d}_i \leq \epsilon_3
        \\
        \texttt{Obstacle},& \text{otherwise}
        \end{cases}
\end{split},
\end{equation}
where $\epsilon_3$ denotes the point-to-plane distance threshold.

\section{Experiments}
To demonstrate our contributions, we conducted quantitative and qualitative comparisons.
For quantitative evaluations, we leveraged various distributed data from single scans to accumulated partial maps from public datasets, which also provide ground-truth semantic labels and poses.
The parameter specifications for our proposed method are outlined in Table~\ref{tab:parems}. 
Additionally, to highlight our contributions, we introduce the dataset from extremely bumpy terrain.
\begin{table}[h!]
    \vspace{-3mm}
    \captionsetup{font=footnotesize}
    \centering
    \caption{Parameter setting for B-TMS. Units of $r^{\mathcal{T}}$, $\epsilon_{3}$, and $\theta^{\mathcal{T}}$ are in $m$, $m$, and degree(\textdegree), respectively.}
    \setlength{\tabcolsep}{2.6pt}
    \setlength\extrarowheight{2.4pt}
    \begin{tabular}{l|c|c|c|c|c|c||c|c|c|c|c|c}
        \toprule[1.0pt]
        \multirow{2}{*}{Param.}  & \multicolumn{6}{c||}{For single scans} & \multicolumn{6}{c}{For partial map}\\
        \cline{2-13}
                                 & $r^{\mathcal{T}}$ & $\theta^{\mathcal{T}}$ & $\sigma^{\mathcal{T}}$ & $\epsilon_{1}$ & $\epsilon_{2}$ & $\epsilon_{3}$ & $r^{\mathcal{T}}$ & $\theta^{\mathcal{T}}$ & $\sigma^{\mathcal{T}}$ & $\epsilon_{1}$ & $\epsilon_{2}$ & $\epsilon_{3}$ \\
        \hline
        Value                    & 4 & 20\textdegree & 10 & 0.03 & 0.1 & 0.125 & 2 & 20\textdegree & 10 & 0.03 & 0.1 & 0.3 \\
        \bottomrule[1.0pt]
    \end{tabular}
    \label{tab:parems}
    \vspace{-4mm}
\end{table}

\subsection{Dataset}
\subsubsection{SemanticKITTI Dataset}
For quantitative comparison on a real-world urban scene dataset, we utilized the SemanticKITTI dataset~\cite{behley2019semantickitti}, which was acquried with Velodyne HDL-64E LiDAR mounted on a vehicle.
It's important to note that the points labeled as \texttt{road}, \texttt{parking}, \texttt{sidewalk}, \texttt{other ground}, \texttt{lane marking}, \texttt{vegetation}, and \texttt{terrain} are considered to be the ground-truth terrain points. 

\subsubsection{Rellis-3D Dataset}
For quantitative evaluation in off-road environments, we utilized the RELLIS-3D dataset~\cite{jiang2021rellis3d}, which was acquired with Ouster OS1-64 and Velodyne Ultra Puck mounted on ClearPath Robotics WARTHOG. 
Specifically, we used the Ouster data, as its location serves as the basis for the provided ground-truth pose data. 
It's essential to note that the points labeled as \texttt{grass}, \texttt{asphalt}, \texttt{log}, \texttt{concrete}, \texttt{mud}, \texttt{puddle}, \texttt{rubble}, and \texttt{bush} are considered as ground-truth terrain points. 

\subsubsection{Extremely Bumpy Terrain Dataset}
\begin{figure}[t!]
    \vspace{2mm}
    \captionsetup{font=footnotesize}
    \centering
    \includegraphics[width=\columnwidth]{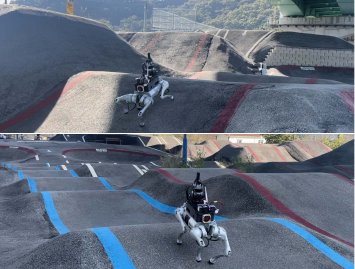}
    \caption{Example scenes of our bumpy terrain dataset, which was acquired by traversing on the curved terrains of various heights and slopes.}
    \label{fig:env_bumpy_terrain}
    \vspace{-5mm}
\end{figure}

To demonstrate the robustness of the proposed method, we acquired our own dataset on the bumpy terrain environments.
As shown in Fig.~\ref{fig:env_bumpy_terrain}, this site covers from slightly to extremely bumpy terrains.
This dataset was acquired using a quadruped robot, specifically the Unitree Go1, equipped with a 3D LiDAR (Ouster OS0-128) and an IMU (Xsens MTI-300).
\vspace{-1mm}
\subsection{Partial Map Generation}
\vspace{-1mm}
To assess segmentation performance on partial maps of various scales, we accumulated scan data with ground-truth labels and voxelized it with $0.2m$ resolution. 
The partial maps were created based on a certain number of sequential frames, with 200 poses for the RELLIS-3D dataset and 500 for the SemanticKITTI dataset.

\subsection{Evaluation Metrics}
Similar to the evaluation methods in our previous studies~\cite{lim21patchwork, oh22travel}, we evaluated terrain segmentation performance using standard metrics: precision (P), recall (R), $F1$-score (F1), and accuracy (A). 
However, there are ambiguous semantic labels such as \texttt{vegetation} of SemanticKITTI and \texttt{bush} of RELLIS-3D cover various plants, which are distinguished differently from terrain. 
To address challenges posed by ambiguous labels such as \texttt{vegetation} and \texttt{bush}, we conducted two evaluations considering the sensor height $h_{s}$: one including the whole data, where only points with $z$-values below $-0.25 \cdot h_{s}$ among the ambiguous labels were considered as ground-truth terrain, and one without these data, excluding the ambiguous labels from the metrics. 

\begin{figure}[t!]
    \captionsetup{font=footnotesize}
    \centering
    \includegraphics[width=\columnwidth]{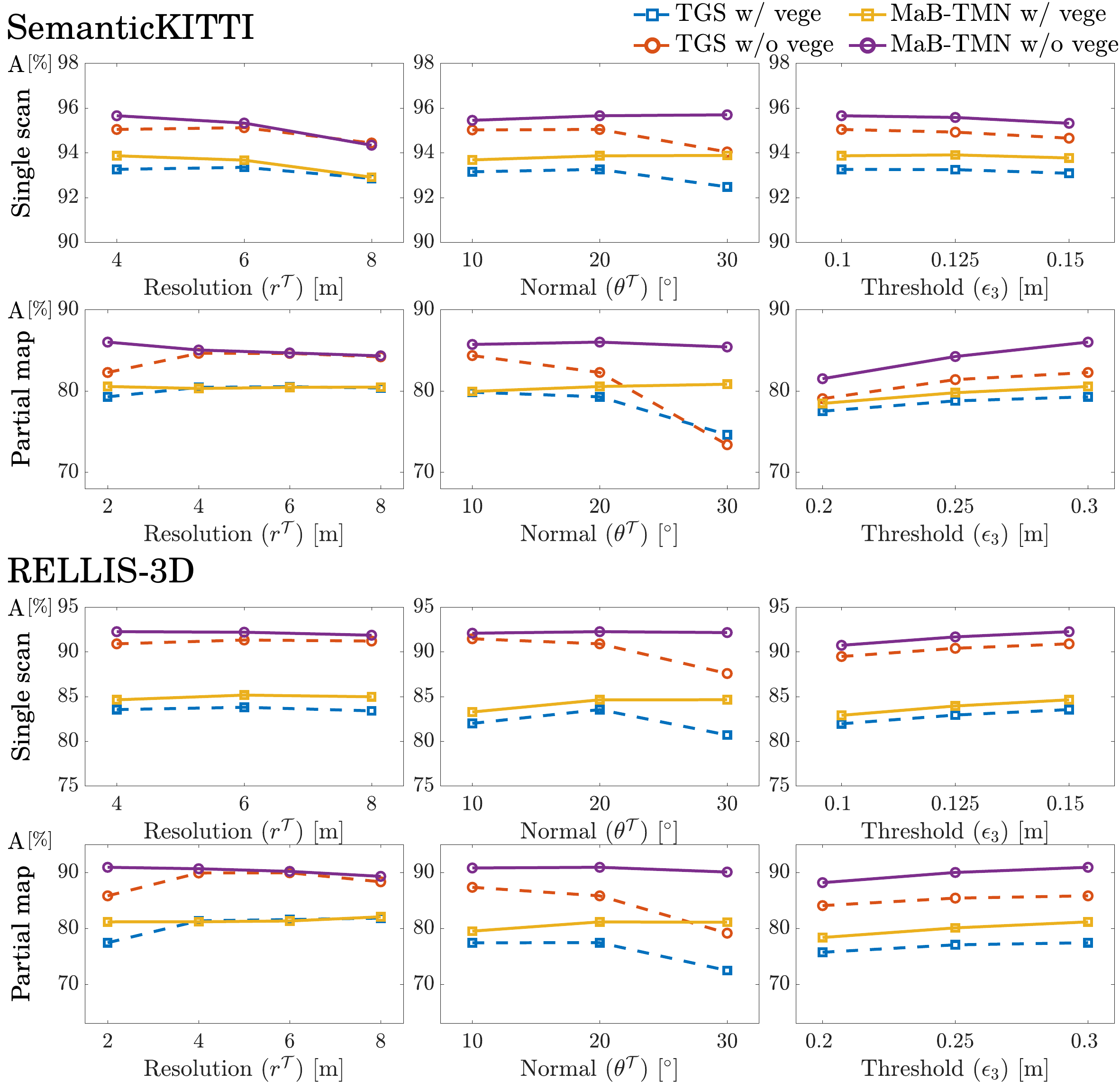}
    \caption{(L-R) The effect of the TGF resolution ($r^{\mathcal{T}}$), the inclination threshold ($\theta^{\mathcal{T}}$), the point-to-plane distance threshold ($\epsilon_{3}$) on ground segmentation for the 3D LiDAR scans and the partial maps from the SemanticKITTI dataset~\cite{behley2019semantickitti}, and RELLIS-3D dataset~\cite{jiang2021rellis3d}.}
    \label{fig:params_study}
    \vspace{-7mm}
\end{figure}

\section{Results and Discussion}

\begin{table*}[t!]
    \captionsetup{font=footnotesize}
    \centering
    \vspace{2mm}
    \caption{
    Quantitative comparison for ground segmentation in terms of computation time (T) measured in [$ms$] and other metrics reported as [\%]. $\mu$ and $\sigma$ represent the mean and standard deviation of each metric, respectively. Computation time results were obtained on an Intel(R) Core i7-8700 CPU. 
    Note that the computation time for partial maps varies depending on the map scale, so the computation time for the partial maps was not measured.
    }
    \vspace{-1mm}
    \setlength{\tabcolsep}{8.4pt}
    \setlength\extrarowheight{0.8pt}
    \begin{tabular}{l||r||c|c|c|c|c|c||c|c|c|c|c|c||c}
        \toprule[1.0pt]
        \multirow{14}{*}{\rotatebox[origin=c]{90}{SemanticKITTI}}&\multirow{3}{*}{Metrics} & \multicolumn{6}{c||}{w/ \texttt{vegetation}} & \multicolumn{6}{c||}{w/o \texttt{vegetation}} & \multirow{2}{*}{T}\\ 
        & & P & R & \multicolumn{2}{c|}{$F_{1}$-score} & \multicolumn{2}{c||}{Accuracy} & P & R & \multicolumn{2}{c|}{$F_{1}$-score} & \multicolumn{2}{c||}{Accuracy}& \\
        && $\mu$ $\uparrow$ & $\mu$ $\uparrow$ & $\mu$ $\uparrow$ & $\sigma$ $\downarrow$ & $\mu$ $\uparrow$ & $\sigma$ $\downarrow$ & $\mu$ $\uparrow$ & $\mu$ $\uparrow$ & $\mu$ $\uparrow$ & $\sigma$ $\downarrow$ & $\mu$ $\uparrow$ & $\sigma$ $\downarrow$ & $\mu$ \\
        \cline{2-15}      
        &\multicolumn{14}{l}{\textbf{Single Scans}} \\ 
        \cline{2-15}
        &RANSAC~\cite{fischler1981ransac}   & 88.2 & 91.3 & 89.0 & 14.7 & 89.8 & 12.4   & 89.9 & 94.0 & 91.3 & 13.4 & 90.5 & 11.5 & 64  \\
        &GPF~\cite{zermas2017fast}          & 91.4 & 83.9 & 85.6 & 18.3 & 88.9 & 12.3   & 94.9 & 77.1 & 81.4 & 25.5 & 82.7 & 19.9 & 20  \\
        &CascadedSeg~\cite{narksri2018slope}& 91.2 & 69.0 & 78.3 & 10.9 & 82.1 & ~5.6   & 95.2 & 74.1 & 83.0 & ~9.6 & 82.2 & ~7.1 & 74  \\ 
        &R-GPF~\cite{lim21erasor}           & 66.2 & 96.0 & 77.1 & 12.2 & 74.0 & 11.4   & 74.7 & \textbf{98.2} & 83.8 & 10.8 & 78.8 & 11.6 & 27  \\
        &Patchwork~\cite{lim21patchwork}    & 92.5 & \textbf{93.8} & 93.0 & \textbf{~3.2} & 93.5 & \textbf{~2.7}   & 94.2 & 97.6 & 95.8 & \textbf{~2.8} & 95.2 & \textbf{~2.8} & 25  \\
        &TRAVEL~\cite{oh22travel}           & \textbf{95.2} & 90.1 & 92.4 & ~3.8 & 93.3 & ~3.1  & \textbf{96.3} & 95.1 & 95.7 & \textbf{~2.8} & 95.0 & \textbf{~2.8} & \textbf{18}  \\
        &B-TMS (Ours)                     & 94.4 & 92.2 & \textbf{93.2} & ~3.9 & \textbf{93.9} & ~3.0   & 95.5 & 97.0 & \textbf{96.2} & ~3.1 & \textbf{95.7} & ~2.9 & 22  \\
        \cline{2-15}   
        &\multicolumn{14}{l}{\textbf{Partial Maps}} \\ 
        \cline{2-15}      
        &TRAVEL~\cite{oh22travel}           & \textbf{93.9} & 65.7 & 76.8 & ~7.7 & 79.2 & ~8.0 & \textbf{96.6} & 77.1 & 85.1  & ~7.7 & 82.3 & ~9.9 & -  \\
        &B-TMS (Ours)                     & 89.9 & \textbf{76.4} & \textbf{82.1} & ~\textbf{6.6} & \textbf{82.6} & \textbf{~7.2} & 93.6 & \textbf{87.0} & \textbf{89.7} & ~\textbf{6.9} & \textbf{86.9} & \textbf{~8.5} & -  \\      
        \midrule[1.0pt]    
        \multirow{11}{*}{\rotatebox[origin=c]{90}{Rellis-3D: Ouster}}&\multicolumn{14}{l}{\textbf{Single Scans}} \\ 
        \cline{2-15}      
        &RANSAC~\cite{fischler1981ransac}   & 71.9 & 96.2 & 81.3 & 12.2 & 75.9 & 10.9 & 82.6 & 95.6 & 87.3 & 12.9 & 85.0 & 10.6 & 18  \\
        &GPF~\cite{zermas2017fast}          & \textbf{96.2} & 65.4 & 76.9 & 12.3 & 77.6 & 10.9 & \textbf{95.4} & 79.8 & 86.1 & 11.3 & 83.8 & 11.2 & 19  \\
        &CascadedSeg~\cite{narksri2018slope}& 63.1 & \textbf{98.3} & 75.1 & 15.2 & 63.3 & 17.2 & 71.1 & \textbf{98.3} & 79.9 & 19.0 & 71.4 & 22.0 & 38  \\ 
        &R-GPF~\cite{lim21patchwork}        & 64.8 & 71.7 & 65.8 & 12.2 & 57.5 & 10.2 & 72.0 & 65.4 & 66.0 & 16.8 & 59.9 & 12.7 & 24  \\
        &Patchwork~\cite{lim21patchwork}    & 87.2 & 81.5 & 83.7 & \textbf{~7.3} & 82.5 & \textbf{~5.0}  & 92.6 & 85.6 & 88.4 & \textbf{~5.0} & 87.5 & ~7.6 & 19  \\
        &TRAVEL~\cite{oh22travel}           & 89.9 & 80.3 & 84.3 & 10.0 & 83.6 & ~6.4 & 94.6 & 89.2 & 91.4 & ~8.4 & 90.9 & \textbf{~6.2} & \textbf{14}  \\
        &B-TMS (Ours)                     & 89.3 & 83.7 & \textbf{85.7} & 10.6 & \textbf{84.6} & ~7.9 & 94.2 & 91.6 & \textbf{92.5} & ~8.5 & \textbf{92.3} & \textbf{~6.2} & 16  \\
        \cline{2-15}    
        &\multicolumn{14}{l}{\textbf{Partial Maps}} \\ 
        \cline{2-15}      
        &TRAVEL~\cite{oh22travel}           & \textbf{84.4} & 71.5 & 76.4   & 10.6 & 80.5 & ~8.6 & \textbf{91.4} & 80.0 & 84.2   & \textbf{11.6} & 87.7 & ~8.8 & -  \\
        &B-TMS (Ours)                     & 80.7 & \textbf{83.9} & \textbf{81.3} & ~\textbf{9.8} & \textbf{83.5} & \textbf{~6.5}& 88.8 & \textbf{90.3} & \textbf{88.3} & 13.1 & \textbf{92.2} & ~\textbf{6.3} & -  \\
        \bottomrule[1.0pt]
        
    \end{tabular}
    \label{tab:quantitative_comparison_gseg}
\end{table*}
\begin{figure*}[t!]
    \vspace{2mm}
    \captionsetup{font=footnotesize}
    \centering
    \includegraphics[width=\textwidth]{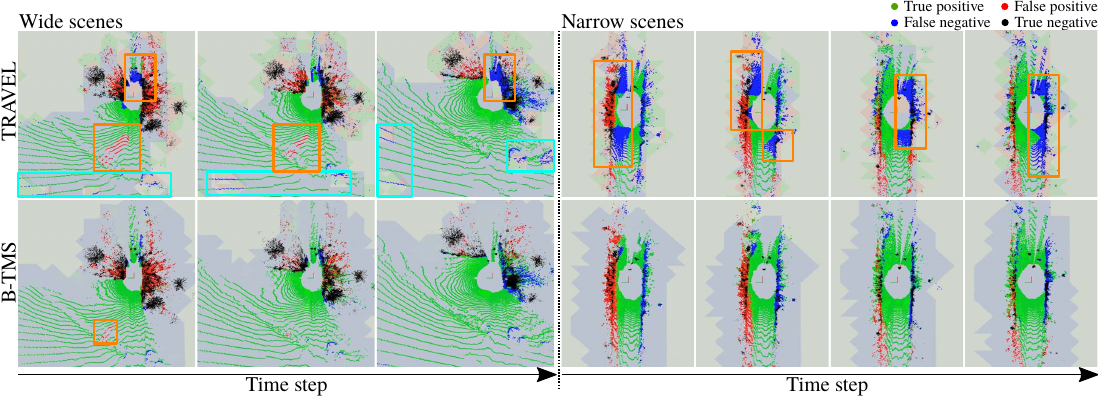}
    \caption{
    Qualitative terrain segmentation results from a sequence of single scans of the RELLIS-3D dataset~\cite{jiang2021rellis3d}, comparing our previous work, TRAVEL~\cite{oh22travel}, with the proposed method, B-TMS. Green, red, blue, and black points represent \textit{true positives}, \textit{false positives}, \textit{false negatives}, and \textit{true negatives}, respectively. 
    B-TMS, employing BGK-based terrain model completion, demonstrates robustness in narrow areas or rough off-road scenes where TRAVEL encounters difficulties, as highlighted by orange boxes. 
    Although single scan data vary in distribution depending on the measured distance, a factor that can limit terrain modeling as highlighted by cyan boxes, B-TMS consistently shows robust results despite these challenges.
    }
    \label{fig:exp_single_scans}
    \vspace{2mm}
    
    \includegraphics[width=\textwidth]{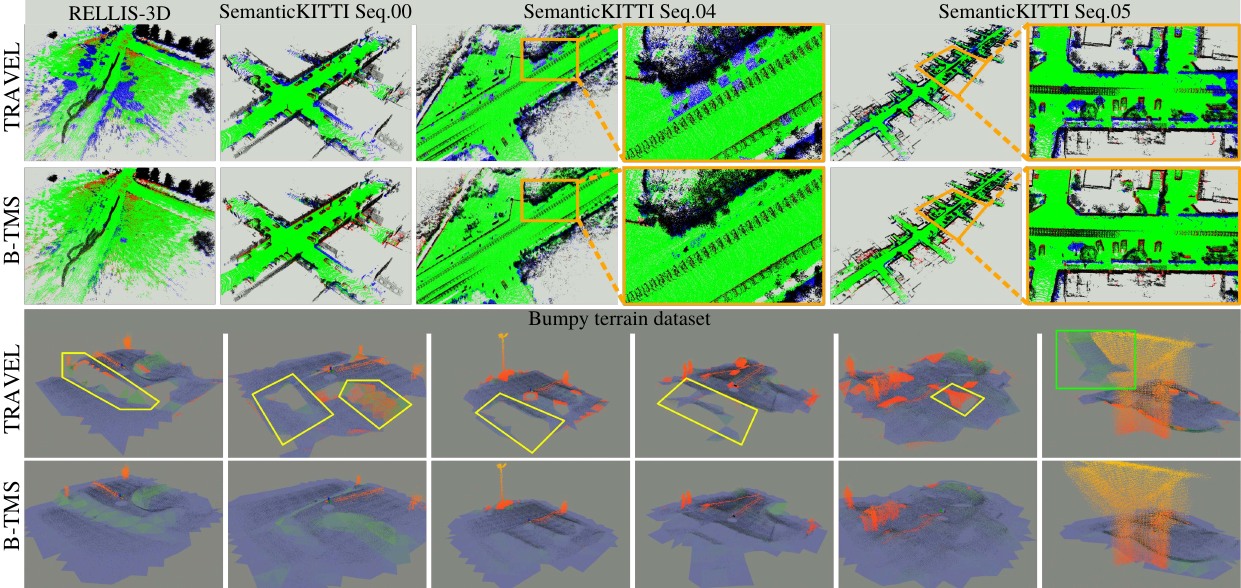}
    \caption{
    Qualitative terrain segmentation results on partial maps of the RELLIS-3D and SemanticKITTI datasets, as well as on local maps from our bumpy terrain dataset, comparing our previous work, TRAVEL~\cite{oh22travel}, and the proposed method, B-TMS.
    In the top two rows, green, red, blue, and black points represent \textit{true positives}, \textit{false positives}, \textit{false negatives}, and \textit{true negatives}, respectively.
    In the bottom two rows, black points indicate estimated terrain, while other points represent obstacles. The green-to-blue plane represents the estimated terrain model with verticality.
    B-TMS, using BGK-based terrain model completion, significantly reduces \textit{false negative} estimates in off-road regions, near walls, and under objects, where TRAVEL encounters challenges. Additionally, the proposed approach mitigates issues arising from the ceiling (green box) and \textit{false negatives} in sunken areas (yellow boxes).
    }
    \label{fig:exp_partial_map}
    \vspace{-2mm}
\end{figure*}

\subsection{Resilience Against Parameter Changes}
We first shed light on the effect of key parameters on terrain segmentation performance, by comparing with our previous work~\cite{oh22travel}.
Fig.~\ref{fig:params_study} illustrates changes in accuracy depending on the TGF resolution ($r^{\mathcal{T}}$), the inclination threshold ($\theta^{\mathcal{T}}$), and the distance threshold ($\epsilon_3$), both with and without considering \texttt{vegetation} and \texttt{bush}.
The two algorithms exhibit similar performance changes in response to $\epsilon_3$ changes. However, for $r^{\mathcal{T}}$ and $\theta^{\mathcal{T}}$, which are used to establish the tri-grid field (TGF), the proposed method demonstrates significantly reduced performance variations compared to TRAVEL. This suggests that BGK-based terrain model completion on TGF addresses problems arising from the inherent limitations of constant resolution and thresholds.

\subsection{Robustness to Data Distribution}
As evident in Table~\ref{tab:quantitative_comparison_gseg} and Figs.~\ref{fig:exp_single_scans} and~\ref{fig:exp_partial_map}, we conducted performance evaluations on single scans, locally accumulated maps, and large-scale partial maps. 
Particularly, Table~\ref{tab:quantitative_comparison_gseg} indicates that, regardless of whether ambiguous labels are considered in the evaluation metrics or not, we achieved the highest $F1$-score and accuracy performance across off-road datasets, urban scene datasets, single scans, and partial maps.
Moreover, as shown in Fig.~\ref{fig:exp_single_scans}, the results of the single scans, which vary in distribution depending on the measured distance, highlights not only the robustness to data distributions, but also the stability on the wide and narrow off-road scenes.
Although the introduction of the BGK-based terrain prediction module slightly increases the computation time compared to our previous work~\cite{oh22travel}, it is nonetheless still suitable for real-time navigation with onboard systems.

\subsection{Adaptability to Diverse Environmental Conditions}
Figs.~\ref{fig:exp_single_scans} and~\ref{fig:exp_partial_map} illustrate qualitative performance comparisons in various environmental conditions. 
A closer look at the top two rows of Fig.~\ref{fig:exp_partial_map} reveals a significant reduction in \textit{false negatives}, previously common in off-road regions, near walls, and under objects.
This reduction aligns with the performance improvements shown in Table~\ref{tab:quantitative_comparison_gseg}. 
Moreover, to assess in diverse terrain environments, we introduced data from extremely bumpy terrain environments. 
The existing approach struggles with terrain modeling failures due to three causes: a) insufficient data in unobservable areas, b) terrain model outliers caused by overhanging objects, resulting \textit{false positives} commonly in off-road scenarios, and c) inappropriate terrain model estimations for bumpy areas, resulting in \textit{false negatives}. 
Our proposed algorithm, featuring BGK-based terrain model prediction and normalized weight-based terrain model fitting, overcomes these outlier issues, enabling stable terrain model predictions.

\section{Conclusion}

In this study, we presented a robust map-wise terrain modeling and segmentation method that combines BGK-based terrain model completion with an efficient graph and node-wise PCA-based traversability-aware terrain segmentation approach. Our results demonstrate the consistent outperformance of B-TMS in the face of parameter variations, changes in data distributions, and alterations in environmental conditions.
Furthermore, we anticipate that the capability to predict terrain models for unobservable and sunken regions will have a positive impact on subsequent autonomous navigation algorithms, particularly contributing to improved navigation performance in off-road scenarios. 

However, despite the robust terrain modeling of our approach, which is based on statistical traversability analyzing the distribution of 3D data, it should also incorporate another method of traversability estimation from semantic information, similar to the approach in the research of Shaban~\textit{et al.}~\cite{shaban22a}, for safer navigation.
In addition, limitations stemming from pose drift along the $z$-axis restrict B-TMS from properly recognizing terrains and evaluating whole maps.
To address these limitations, we will focus on expanding the approach with a terrain-aware loop-closure module to enhance pose estimation performance based on the research of Lim~\textit{et al.}~\cite{lim2023quatro++}, and extend it to whole map-based terrain recognition techniques.

\bibliographystyle{IEEEtran}
\bibliography{./main}

\end{document}